\documentclass{article}[11pt]

\usepackage{amsfonts,amsmath,amscd,dsfont,mathrsfs,amsthm}
\usepackage{graphicx,subcaption,caption}
\usepackage{fullpage}

\newcommand{\figdir}{.}
\newcommand{\bibdir}{.}

\newcommand{\matsnorm}[2]{|\!|\!| #1 | \! | \!|_{{#2}}}
\newcommand{\vecnorm}[2]{\| #1\|_{#2}}

\newcommand{\tracer}[2]{\ensuremath{\langle \!\langle {#1}, \; {#2}
\rangle \!\rangle}}
\newcommand{\opnorm}[1]{\ensuremath{\matsnorm{#1}{2}}}
\newcommand{\frob}[1]{\ensuremath{\matsnorm{#1}{F}}}
\newcommand{\frobnorm}[1]{\ensuremath{\frob{#1}}}
\newcommand{\nuclear}[1]{\ensuremath{\matsnorm{#1}{\operatorname{\tiny{nuc}}}}}
\newcommand{\nucnorm}[1]{\ensuremath{\nuclear{#1}}}
\newcommand{\infnorm}[1]{\vecnorm{#1}{\infty}}
\DeclareMathOperator{\argmin}{argmin}
\DeclareMathOperator{\trace}{trace}

\newcommand{\prob}{\mathbb{P}}
\newcommand{\E}{\mathbb{E}}

\newcommand{\reals}{\mathbb{R}}

\theoremstyle{plain}

\newtheorem{theo}{Theorem}[section]

\newtheorem{lem}{Lemma}[section]
\newtheorem{prop}{Proposition}[section]
\newtheorem{cor}{Corollary}[section]

\theoremstyle{definition} 

\newtheorem{nota}{Notation}[section]
\newtheorem{de}{Definition}[section]
\newtheorem{exa}{Example}[section]
\newtheorem{as}{Assumption}[section]
\newtheorem{alg}{Algorithm}[section]

\newcommand{\btheo}{\begin{theo}}
\newcommand{\bde}{\begin{de}}
\newcommand{\ble}{\begin{lem}}
\newcommand{\bpr}{\begin{prop}}
\newcommand{\bno}{\begin{nota}}
\newcommand{\bex}{\begin{exa}}
\newcommand{\bcor}{\begin{cor}}
\newcommand{\spro}{\begin{proof}}
\newcommand{\bas}{\begin{as}}
\newcommand{\balg}{\begin{alg}}

\newcommand{\etheo}{\end{theo}}
\newcommand{\ede}{\end{de}}
\newcommand{\ele}{\end{lem}}
\newcommand{\epr}{\end{prop}}
\newcommand{\eno}{\end{nota}}
\newcommand{\eex}{\end{exa}}
\newcommand{\ecor}{\end{cor}}
\newcommand{\fpro}{\end{proof}}
\newcommand{\eas}{\end{as}}
\newcommand{\ealg}{\end{alg}}

\theoremstyle{plain}

\newtheorem{theos}{Theorem}
\newtheorem{props}{Proposition}
\newtheorem{lems}{Lemma}
\newtheorem{cors}{Corollary}

\theoremstyle{definition}
\newtheorem{exas}{Example}
\newtheorem{algs}{Algorithm}
\newtheorem{asss}{Assumption}
\newtheorem{defns}{Definition}

\newcommand{\btheos}{\begin{theos}}
\newcommand{\etheos}{\end{theos}}
\newcommand{\bprops}{\begin{props}}
\newcommand{\eprops}{\end{props}}
\newcommand{\bdes}{\begin{defns}}
\newcommand{\edes}{\end{defns}}
\newcommand{\blems}{\begin{lems}}
\newcommand{\elems}{\end{lems}}
\newcommand{\bcors}{\begin{cors}}
\newcommand{\ecors}{\end{cors}}
\newcommand{\bexs}{\begin{exas}}
\newcommand{\eexs}{\end{exas}}
\newcommand{\balgs}{\begin{algs}}
\newcommand{\ealgs}{\end{algs}}
\newcommand{\bass}{\begin{asss}}
\newcommand{\eass}{\end{asss}}

\newcommand{\dimd}{d}
\newcommand{\dima}{\dimd_1}
\newcommand{\dimb}{\dimd_2}
\newcommand{\rdim}{r}

\newcommand{\Param}{\Theta}
\newcommand{\Thetastar}{\Param^*}
\newcommand{\Thetahat}{\widehat{\Param}}
\newcommand{\rspace}{\Omega}
\newcommand{\personex}{p}

\newcommand{\numobs}{n}
\newcommand{\obs}[1]{y_{#1}}
\newcommand{\design}[1]{X^{(#1)}}

\newcommand{\Loss}{\mathcal{L}_{\numobs}}
\newcommand{\itema}[1]{l(#1)}
\newcommand{\itemb}[1]{j(#1)}
\newcommand{\useri}[1]{k(#1)}
\newcommand{\stand}[1]{e^{(#1)}}
\newcommand{\spiky}{\alpha}
\newcommand{\mylogit}[1]{\frac{\exp(#1)}{1+\exp(#1)}}
\newcommand{\svds}[1]{\sigma_{#1}}
\newcommand{\comment}[1]{}

\newcommand{\setA}{\mathcal{A}}
\newcommand{\1}{\mathbf{1}}
\newcommand{\noise}{\xi}

\begin{document}
\begin{center}
  
  {\bf{\LARGE{Individualized Rank Aggregation using Nuclear Norm Regularization}}}
  
  \vspace*{.2in}

  {\large{
      \begin{tabular}{ccc}
        Yu Lu & & Sahand N. Negahban\\
      \end{tabular}
    }}
  
  \vspace*{.2in}
  
  \begin{tabular}{c}
    Department of Statistics\\
    Yale University
  \end{tabular}
  
  \vspace*{.2in}

  \today
  
  \vspace*{.2in}
  
  \begin{tabular}{c}
    Technical Report \\
    Department of Statistics,  Yale University
  \end{tabular}
\begin{abstract}
  In recent years rank aggregation has received significant attention
  from the machine learning community.  The goal of such a problem is
  to combine the (partially revealed) preferences over objects of a
  large population into a single, relatively consistent ordering of
  those objects. However, in many cases, we might not want a single
  ranking and instead opt for individual rankings. We study a version
  of the problem known as collaborative ranking. In this problem we
  assume that individual users provide us with pairwise preferences
  (for example purchasing one item over another). From those
  preferences we wish to obtain rankings on items that the users have
  not had an opportunity to explore. The results here have a very
  interesting connection to the standard matrix completion problem. We
  provide a theoretical justification for a nuclear norm regularized
  optimization procedure, and provide high-dimensional scaling results
  that show how the error in estimating user preferences behaves as
  the number of observations increase.  \comment{ We complement the
    theoretical analysis with an empirical study demonstrating the
    validity of the results.  }
\end{abstract}
\end{center}
\section{Introduction}
We have seen a number of recent advancements to the theory of rank
aggregation. This problem has a number of applications ranging from
marketing and advertisements to competitions and election. The main
question of rank aggregation is how to consistently combine various
individual preferences. This type of data is frequently available to
us: what webpage did a user select, who won the chess match, which
movie did a user watch, etc.... All of these examples yield
comparisons without explicitly revealing an underlying score. That is,
only the preference is observed, not necessarily the strength of the
preference (in the case of sports one might argue that the score
indicates such a magnitude difference). Additionally, numeric scores
have been shown to be inconsistent and subject to variations in
calibration in various contexts. Given how natural the problem of rank
aggregation is, there has been a wide body
recent~\cite{Duchi10,ammarandshah} and classical
work~\cite{Arrow,Bradley,Condorcet,Luce} to understand how to
consistently combine preferences. However, all of these methods have a
major drawback: they aim to find \emph{one} ranking. In many settings,
various individuals will have separate preferences, and we wish to
model those distinctions. For example, we might wish to provide
personalized ads, search results, or movie recommendations on a per
user basis. In standard contexts we assume that there is one
consistent ranking that does well to approximate the behavior of all
users, but these aggregation methods cannot model the discrepancies
across users. Our goal is to understand how to analyze a method that
has the flexibility to account for user differences and can be
adaptive; that is, if there are no differences, then the method should
have stronger performance guarantees. This task can be seen as rank
aggregation analog to the standard collaborative filtering problem.

\comment{ The matrix completion problem has received extensive
  extension over the past decade and recently a number of theoretical
  advances have been made in understanding the matrix completion
  problem. Recent work has aimed to generalize the types of
  observation models that we can considering in the matrix completion
  framework. One popular framework has been the so-called ``one-bit''
  matrix completion problem~\cite{onebit} where we only observe a $+1$
  or $-1$ at each entry. This model is also the effective one
  considered by Srebro et. al.~\cite{srebro} in much of their early
  work on matrix completion. Matrix completion can also be seen as an
  approach to the collaborative filtering
  problem~\cite{collabfiltering}.}

While there have been significant theoretical advances in the
understanding of collaborative filtering, or more generally matrix
completion~\cite{CanRecCompletion,TsyCompletion,NegWaiCompletion},
there has been far less work in understanding how to perform
the proposed type of collaborative ranking. Recent work has
demonstrated that taking rankings into consideration can significantly
improve upon rating prediction
accuracy~\cite{eigenrank,param17,cofirank,Yi13}, thus it is a natural
question to understand how such collaborative ranking methods might
behave. One reason for this discrepancy is this theoretical
understanding of single user rank aggregation is already a
very challenging problem as discussed above. Whereas, single rating
aggregation is trivial: take an average. Another, possibly more
interesting distinction is in the amount of apparent information made
available. In the standard matrix completion setting we have direct
(albeit noisy) access to the true underlying ratings. Therefore, if
the noise is sufficiently small, we could order the information into a
list. On the other hand, in the collaborative ranking problem we never
have direct access to the true signal itself and only observe relative
differences. In some sense, this is a harder problem~\cite{shahetal14}
owing to the fact that the comparisons are in themselves functions of
the underlying ratings. When we are given, for example, $\personex$
ratings, then we can convert that to $\binom{\personex}{2}$ pairwise
comparisons. This crude analysis seems to indicate that we would
require far greater pairwise comparisons in order to recover the true
underlying matrix. We will show that this increase in the number of
examples is not required. In the sequel, we will show that under a
natural choice model for collaborative ranking, the total number of
comparisons needed to estimate the parameters is on the same order as
the total number of explicit ratings observations required in the
standard matrix completion literature. Thus, we demonstrate that
collaborative ranking based pair-wise comparisons from a simple and
natural model can yield very similar results as in the standard matrix
completion setting.

\paragraph{Past Work} As alluded to above there has been some work in
understanding collaborative rankings and learning user preferences.
The nuclear norm approach is fundamentally a regularized
$M$-estimator~\cite{Neg09}.  The application of the nuclear norm
approach to collaborative ranking was first proposed by Yi et
al.~\cite{Yi13}. There work showed very good empirical evidence for
using such a nuclear norm regularized based approach. However, that
work left open the question of theoretical guarantees. Other results
also assume that the underlying ratings are in fact
available. However, rather than inferring unknown ratings their goal
is to infer unknown ranked preferences \emph{from} known ratings. That
is, they wish to deduce if a user will prefer one item over another
rather than guess what their ratings of that item might
be~\cite{eigenrank,cofirank,bayesmatrix}. The work by by Weimer
et. al.~\cite{cofirank} also uses a nuclear norm regularization, but
that work assumes access to the true underlying
ratings, while we assume access only to pairwise preferences. Other
algorithms aggregate users’ ratings by exploiting the similarity of
users by nearest neighbor search~\cite{empirical,param17}, low-rank
matrix factorization ~\cite{matrixfact05,bayesmatrix,cofirank}, or
probabilistic latent model \cite{latent04,problatent}.  However, as
noted, numeric ratings can be highly varied even when preferences are
shared.

Pairwise preference based ranking methods can effectively address the
limitations of rating based methods. Furthermore, numerical ratings
can always be transformed into pairwise comparisons. Salimans et
al.~\cite{salimans2012} use a bilinear model and do estimation in the
Bayesian framework.  Liu et al.~\cite{problatent} use the
Bradley-Terry-Luce (BTL) Model. Rather than our low-rank setting, they
characterize the similarity between different users by using a mixture
model. Both methods are computationally inefficient. More important,
all these methods fail to provide theoretical justifications of their
algorithms.

There are some theoretical works for learning a single ranking list
from pairwise comparisons. Work by Jamieson and Nowak~\cite{JamNow11}
seeks to exploit comparisons to significantly reduce the number of
samples required to obtain a good estimate of an individual's utility
function. Their method demonstrates that when the objects exist in a
lower-dimensional space, then the number of queries required to learn
the user's utility significantly decreases. One drawback of their
approach is that the authors must assume that descriptors or features
for the underlying objects are provided; which is not necessarily the
case in all contexts. Negahban et al. \cite{NegOhSha12} propose the
Rank Centrality algorithm and show rate optimal (up to log factors)
error bounds of their algorithm under BTL model. They also provide
theoretical analysis of penalized maximum likelihood estimator, which
serves as an inspiration of our work.

\paragraph{Our contributions} In this report, we present the first
theoretical analysis of a collaborative ranking algorithm under a
natural observation model. The algorithm itself is quite simple and
falls into the framework of regularized $M$-estimator~\cite{Neg09}. We
provide finite sample guarantees that hold with high probability on
recovering the underlying preference matrix. Furthermore, the
techniques outlined in the proof section our general and can be
applied to a variety of sampling operators for matrix completion. For
example, a simple modification of our proof yields a different class
of results for the ``one-bit'' matrix completion
problem~\cite{onebit}.

In the following we present an explicit description of our model in
Section~\ref{sec:model}. In Section~\ref{sec:estimate} we present the
proposed estimation procedure that we wish to analyze. Finally, in
Section~\ref{sec:main} we provide a statement of the main theorem
followed by experiments in Section~\ref{sec:experiments}. Finally, in
Section~\ref{sec:proofs} we present the proof.

\comment{ %%%%%%comment 420
  Nevertheless, those methods demonstrate that while people's ratings
  might be inconsistent, their preferences (e.g. preferring item $a$
  over item $b$) remain well behaved. There has been growing recent
  attention in understanding how to aggregate partial rankings from
  various users in a consistent manner. While there has been some
  theoretical advancements in this area, most of the work has focused
  on aggregating results into single rankings.  This generalization of
  collaborative filtering yields the problem of collaborative ranking.
  This context is natural as in many settings a user might pick one
  items versus others: for example at the super market, when selecting
  a movie among a choice of movies, or when picking a webpage based on
  her search query. Other collaborative ranking approaches have proven
  to be very promising~\cite{param17,cofirank}, but theoretical
  justifications for their use have not been as explored while the
  standard matrix completion
  problem~\cite{CanRecCompletionCanRecCompletion} has received a
  significant amount of theoretical
  attention.} %%%%%%%% end comment 420

\paragraph{Notation:}
For a positive integer $n$ we will let $[n] = \{1,2,\hdots,n\}$ be the
set of integers from $1$ to $n$. For two matrices $A$, $B \in
\reals^{\dima \times \dimb}$ of commensurate dimensions, let
$\tracer{A}{B} = \trace(A^T B)$ be the trace inner product. For a
matrix $A \in \reals^{\dima \times \dimb}$ let $A_{i,j}$ denote the
entry in the $i^{th}$ row and $j^{th}$ column of $A$. Take
$\svds{i}(A)$ to be the $i^{th}$ singular value of $A$ where
$\svds{i}(A) \geq \svds{i+1}(A)$. Let $\opnorm{A} = \svds{1}(A)$,
$\nucnorm{A} = \sum_{j=1}^{\min(\dima,\dimb)} \svds{j}(A)$ be the
nuclear norm of $A$, i.e. the sum of the singular values of $A$, and
$\frobnorm{A} =
\sqrt{\tracer{A}{A}}=\sqrt{\sum_{j=1}^{\min(\dima,\dimb)}
  \svds{j}^2(A)}$ to be the Frobenius norm of $A$. Finally, we let
$\infnorm{A} = \max_{i,j} |A_{i,j}|$ to be the elementwise infinity
norm of the matrix $A$.

\section{Problem Statement and Model}
\label{sec:model}
In this section we provide a precise description of the underlying
statistical model as well as our problem.
\subsection{Data and Observation Model}
Recall that each user provides a collection of pairwise preferences
for various items. We assume that the data are the form
$(\design{i},\obs{i})$ where $\design{i} \in \reals^{\dima \times
  \dimb}$. We assume that the $i^{th}$ piece of data is a query to
user $\useri{i}$ asking if she prefers item $\itema{i}$ to item
$\itemb{i}$. If she does, then $\obs{i}=1$, otherwise $\obs{i}=0$. In
other words, $\obs{i} = 1$ if user $\useri{i}$ prefers item
$\itema{i}$ to item $\itemb{i}$, otherwise $\obs{i}=0$. Let the
underlying (unknown and unobservable) user preferences be encoded in
the matrix $\Thetastar \in \reals^{\dima \times \dimb}$ such that
$\Thetastar_{k,j}$ is the score that user $k$ places on item $j$. We
will also assume that $\frobnorm{\Thetastar} \leq 1$ to normalize the
signal. For identifiability we assume that the sum of the rows of
$\Thetastar$ is equal to zero. We must also assume that
$\infnorm{\Thetastar} \leq \frac{\spiky}{\sqrt{\dima \dimb}}$. Similar
assumptions are made in the matrix completion literature and is known
to control the ``spikyness'' of the matrix. Both of these assumptions
are discussed in the sequel. For compactness in notation we let
$\design{i} = \sqrt{\dima \dimb} \stand{\useri{i}} (\stand{\itema{i}}
- \stand{\itemb{i}})^T$ where $\stand{a}$ is the standard basis vector
that takes on the value $1$ in the $a^{th}$ entry and zeros everywhere
else. Taking the trace inner product between $\Thetastar$ and
$\design{i}$ yields
\begin{equation*}
  \tracer{\Thetastar}{\design{i}} = \sqrt{\dima \dimb} \ ( \Thetastar_{\useri{i},\itema{i}} - \Thetastar_{\useri{i},\itemb{i}} \ )
\end{equation*}
and denotes the relative preference that user $\useri{i}$ has for item $\itema{i}$
versus $\itemb{i}$. Our observation model takes the form
\begin{equation}
  \label{eq:model}
  \prob(\obs{i}=1 | \itema{i}=l, \itemb{i}=j, \useri{i}=k) = \mylogit{\tracer{\Thetastar}{\design{i}}}
\end{equation}
The above is the standard Bradley-Terry-Luce model for pairwise
comparisons. In full generality, one can also consider the Thurstone
models for pairwise preferences.

We shall take $\Thetastar$ to be low-rank or well approximate by a
low-rank matrix. This is analogous to the matrix completion literature
and models the fact that the underlying preferences are derived from
latent low-dimensional factors. In this way, we can extract features
on items and users without explicit domain knowledge.
\paragraph{Discussion of assumptions:}
In the above we assume that the $\ell_\infty$ norm of the matrix is
bounded. This form of assumption is required for estimating the
underlying parameters of the matrix and can be thought of as an
incoherence requirement in order to ensure that the matrix itself is
not orthogonal to the observation operator. For example, suppose that
we have a matrix that is zeros everywhere except in one row where we
have a single $+1$ and a single $-1$. In that case, we would never be
able to recover those values from random samples without observing the
entire matrix. Hence, the error bounds that we derive will include
some dependency on the infinity norm of the matrix. If generalization
error bounds are the desired outcome, then such requirements can be
relaxed at the expense of slower error convergence guarantees and no
guarantees on individual parameter recovery. Also noted above is the
requirement that the sum of each of the rows of $\Thetastar$ must be
equal to $0$. This assumption is natural owing to the fact that we can
ever only observe the differences between the intrinsic item
ratings. Hence, even if we could exactly observe all of those
difference, the solution would not be unique up to linear offsets of
each of the rows. We refer the reader to other work in matrix
completion~\cite{CanRecCompletion,NegWaiCompletion} for a discussion
of incoherence.

\section{Estimation Procedure}
\label{sec:estimate}
We consider the following simple estimator for performing
collaborating ranking. It is an example of a regularized
$M$-estimator~\cite{Neg09}.
\begin{equation}
\label{estprocedure}
  \Thetahat = \argmin_{\Param \in \rspace} \underbrace{\frac{1}{n} \sum_{i=1}^n \log(1+\exp(\tracer{\Param}{\design{i}})) - \obs{i} \tracer{\Param}{\design{i}}}_{\Loss(\Param)} + \lambda \nucnorm{\Param},
\end{equation}
where $\Loss(\Param)$ is the random loss function and
\begin{equation*}
  \Omega = \{A \in \reals^{\dima \times \dimb} \mid \infnorm{A} \leq \spiky, \text{ and  $\forall j \in [\dima]$ we have $\sum_{k=1}^{\dimb} A_{j,k} = 0$} \}
\end{equation*}
This method is a convex optimization procedure, and very much related
to the matrix completion problems studied in the literature. A few
things to note about the constraint set presented above. While in
practice, we do not impose the $\ell_\infty$ constraint, the theory
requires us to impose the condition and an interesting line of work
would be to remove such a constraint. A similar constraint appears in
other matrix completion work~\cite{NegWaiCompletion}. As discussed
above, the second condition is a fundamental one. It is required to guarantee
identifiability in the problem even if infinite data were available.

The method itself has a very simple interpretation. The random loss
function encourages the recovered parameters to match the
observations. That is, if $y_i = 1$ then we expect that
$\Thetastar_{\useri{i},\itema{i}} >
\Thetastar_{\useri{i},\itemb{i}}$. The second term is the nuclear
norm and that encourages the underlying matrix $\Thetastar$ to be
low-rank~\cite{CanRecCompletion}.

\section{Main Results} \label{sec:main}
In this section we present the main results of our paper, which
demonstrates that we are able to recover the underlying parameters
with very few total observations. The result is analogous to similar
results presented for matrix completion~\cite{TsyCompletion,SewoongCompletion, NegWaiCompletion},
\begin{theos}
  \label{maintheorem}
  Under the described sampling model, let $d=(\dima + \dimb)/2$, assume $\numobs < d^2 \log d$, and take $\lambda \geq 32 \sqrt{\frac{\dimd \log
      \dimd}{\numobs}}$. Then, we have that the Frobenius norm of the
  error $\Delta = \Thetahat - \Thetastar$ satisfies
  \begin{equation*}
    \frobnorm{\Delta} \le c_1 \max \left ( \spiky,\frac{1}{\psi(2 \alpha)} \right ) \max \left \{\sqrt{\frac{r\dimd \log \dimd}{n}}, \left( \sqrt{\frac{r\dimd \log \dimd}{n}}\sum_{j=r+1}^{\min\{\dima, \dimb\}} \sigma_j(\Thetastar) \right )^{1/2} \right \}
  \end{equation*}
 with probability at least $1 - \frac{2}{d^2}$ for some universal constant $c_1$.
\end{theos}
The above result demonstrates that we can obtain consistent estimates
of the parameters $\Thetastar$ using the convex program outlined in
the previous section. Furthermore, the error bound behaves as a
parametric error rate, that is the error decays as
$\frac{1}{\numobs}$. The result also decomposes into two terms. The
first is the penalty for estimating a rank $\rdim$ matrix and the
second is the price we pay for estimating an approximately low-rank
matrix $\Thetastar$ with a rank $r$ matrix. These results exactly
match analogous results in the matrix completion literature barring
one difference: there is also a dependency on the function $\psi$.
However, this necessity is quite natural since if we are interested in
parameter recovery, then it would be impossible to distinguish between
extremely large parameters. Indeed, this observation is related to
the problem of trying to measure the probability of a coin coming up
heads when that probability is extremely close to one. Other results
in matrix completion also discuss such a requirement as well as the
influence of the spikyness parameter~\cite{TsyCompletion,onebit}.
The proof of this result, for which we provide an outline in Section~\ref{sec:proofs},
follows similar lines as other results for matrix completion.

\section{Experiments}
\label{sec:experiments}
Here we present simulation results to demonstrate the accuracy of the
error rate behavior predicted by Theorem~\ref{maintheorem}.  To
make the results more clean, we consider the exact low rank case here,
which means each individual user's preference vector is the linear
combination of $r$ preference vectors. Then according to our main
results, the empirical squared Frobenius norm error
$\frobnorm{\Thetahat-\Thetastar}^2$ under our estimation procedure~\eqref{estprocedure} will be scaled as $\frac{rd\log d}{n}$. For all
the experiments, we solved the convex program~\eqref{estprocedure} by
using proximal gradient descent with step-sizes from~\cite{AgaNegWai}
for fast convergence via our own implementation in R.
\begin{figure}[h]
\centering
\begin{subfigure}{.5\textwidth}
  \centering
  \includegraphics[width=\linewidth]{\figdir/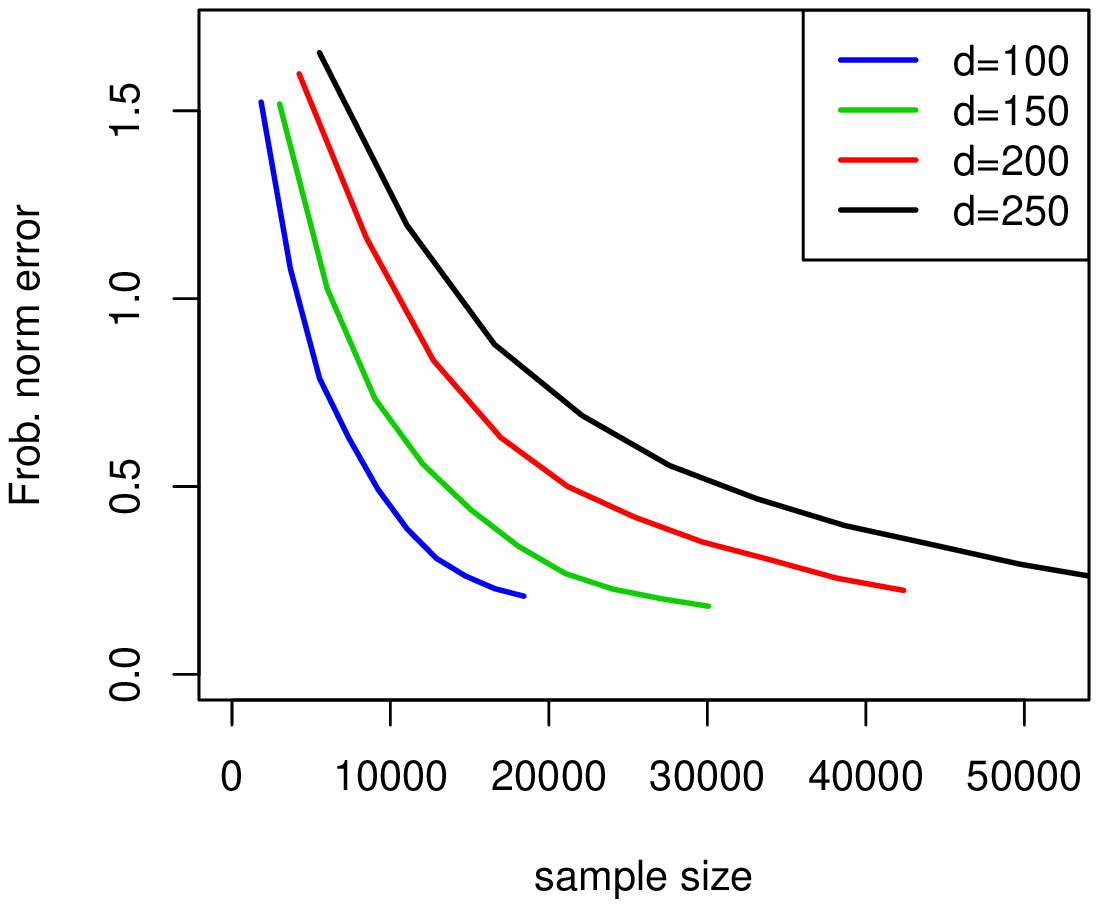}
  \caption{}
  \label{fig:sub1}
\end{subfigure}%
\begin{subfigure}{.5\textwidth}
  \centering
  \includegraphics[width=\linewidth]{\figdir/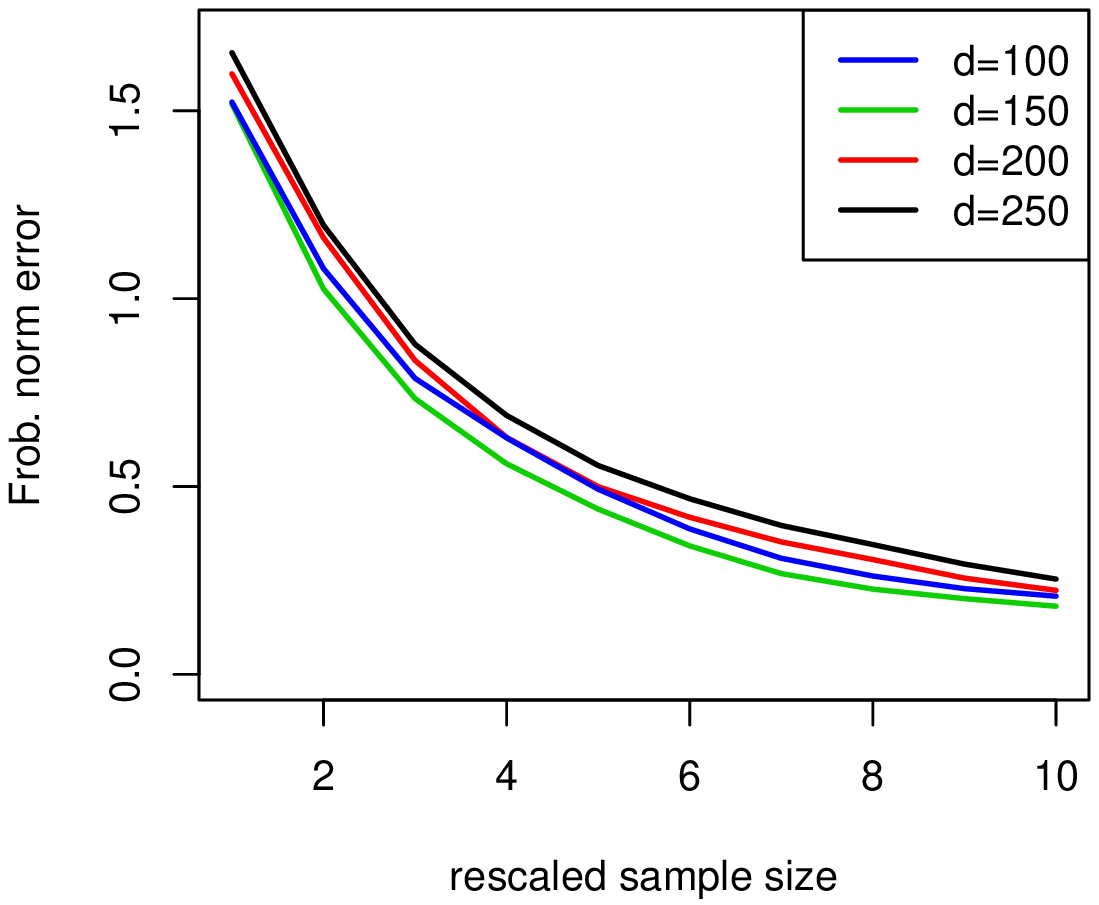}
  \caption{}
  \label{fig:sub2}
\end{subfigure}
\caption{\noindent Plots of squared Frobenius norm error $\frobnorm{\Thetahat-\Thetastar}^2$  when applying estimation procedure (\ref{estprocedure}) on the exact low rank matrix. Each curve corresponds to a different problem size $\dima=\dimb=d \in \{100,150,200,250\}$ with a fixed rank $r=4$. (a) Plots of Frobenius norm error against the raw sample size. As sample size increases, the error goes to zero. (b) Plots of the same Frobenius norm error against rescaled sample size $n/(rd\log d)$, all plots are aligned fairly well as expected by our theory.}
\label{fig:simulation}
\end{figure}

In Figure 1 we report the results of four different problem sizes with
equal user size $\dima$ and item size $\dimb$ and the fixed rank $r$,
where $\dima=\dimb=d \in \{100,150,200,250\}$, $r=4$.  For a given
sample size d, we ran $T=10$ trials and computed the squared Frobenius
norm error $\frobnorm{\Thetahat-\Thetastar}^2$ averaged over those
trials. Panel (a) shows the plots of Frobenius norm error versus raw
sample size. It shows the consistency of our estimation procedure
because the Frobenius norm error goes to zero as sample size
increases. And the curves shift to right as the problem dimension $d$
increases, matching with the intuition that larger matrices require
more samples. In panel (b), we plot the simulation results versus the
rescaled sample size $N=n/(rd\log d)$. Consistent with the prediction
of Theorem \ref{maintheorem}, the error plots are aligned fairly well
and decay at the rate of $1/N$

\section{Proof of Main Result}
\label{sec:proofs}
We now present a proof of the main result. We will use the machinery
developed by Negahban and Wainwright~\cite{NegWaiCompletion} and establish a
Restricted Strong Convexity (RSC) for our loss. The proof follows
standard techniques, with some care when handling the new observation
operator.
\subsection{Proof of Theorem~\ref{maintheorem}}
The key to establishing the RSC condition is to
demonstrate that the error in the first order Taylor approximation of
the loss is lower-bounded by some quadratic function. To that end we
note that for $\Delta = \Param - \Thetastar$ and by the Taylor
expansion we have that
\begin{equation}
  \label{eq:secondordererror}
  \Loss(\Param) - \Loss(\Thetastar) - \tracer{\nabla \Loss(\Thetastar)}{\Delta} = \frac{1}{2 \numobs} \sum_{i=1}^\numobs \psi \left ( \tracer{\Thetastar}{\design{i}} + s \tracer{\Delta}{\design{i}} \right ) \left (\tracer{\Delta}{\design{i}} \right )^2,
\end{equation}
where $s \in [0,1]$ and
\begin{equation*}
  \psi(x) = \frac{\exp(x)}{(1+\exp(x))^2}.
\end{equation*}
Now, we may apply the fact that both $\infnorm{\Thetahat}$,
$\infnorm{\Thetastar} \leq \spiky/\sqrt{\dima \dimb}$ and that
$\psi(x)$ is symmetric and decreases as $x$ increases to obtain that
equation~\eqref{eq:secondordererror} is lower-bounded by:
\begin{equation}
  \label{eq:lowerbound}
  \frac{1}{2 \numobs} \sum_{i=1}^\numobs \psi \left( 2 \spiky \right ) \left (\tracer{\Delta}{\design{i}} \right )^2
\end{equation}
Therefore, it suffices to prove a lower-bound on $\frac{1}{2 \numobs} \left (\tracer{\Delta}{\design{i}} \right )^2$
for all possible vectors $\Delta$. For
that, we present the following lemma.

\begin{lems}
  \label{RSC}
  For $\infnorm{\Theta} \le r_3 := \frac{2\spiky}{\sqrt{\dima \dimb}}$,
  $d=(\dima + \dimb) /2$, and $\numobs < \dimd^2 \log \dimd$. When
  $\design{i}$ are i.i.d observations we have with probability greater than $1-2d^{-2^{18}}$
  \begin{equation*}
    \frac{1}{\numobs} \sum_{i=1}^\numobs \left (\tracer{\Theta}{\design{i}} \right )^2 \ge \frac{1}{3}\frobnorm{\Theta}^2 ~~\text{~~for all $\Theta$ in $\setA$}
  \end{equation*}
  where
    \begin{equation*}
     \mathcal{A} = \left \{\Theta \in \reals^{\dima \times \dimb} \mid 
\infnorm{\Theta} \leq r_3, \frobnorm{\Theta}^2 \ge  128 \spiky \sqrt{\frac{d \log d}{n}}\nucnorm{\Theta} 
\text{ and $\forall j \in [\dima]$ we have $\sum_{k=1}^{\dimb} \Theta_{j,k} = 0$} \right \}
    \end{equation*}
\end{lems}

Another key element for establishing the error is the following upper-bound on the operator norm of a random matrix.
\begin{lems} \label{Lem:crosstermbound} Consider the sampling model
  described above. Then for i.i.d. $(\noise_i,\design{i})$, where
  $|\noise_i| \leq \gamma$ and $\E[\xi_i | \design{i}] = 0$ we have that
\begin{equation*}
  \prob \left ( \opnorm{\frac{1}{\numobs} \sum_{i=1}^\numobs \noise_i \design{i}} > 8\gamma \sqrt{\frac{ \dimd \log \dimd}{\numobs}} \right ) \le \frac{2}{\dimd^2},
\end{equation*} 
\end{lems}

We these two ingredients in hand we may now prove the main result. The
steps are a slight modification of the ones taken for standard matrix
completion~\cite{NegWai11b}. By the optimality of $\Thetahat$ we have
\begin{equation*}
  \Loss (\Thetahat) + \lambda \nucnorm{\Thetahat} \le \Loss (\Thetastar) + \lambda \nucnorm{\Thetastar}
\end{equation*}
Let $\Delta = \Thetahat - \Thetastar$, then
\begin{equation*}
\Loss(\Thetahat) - \Loss(\Thetastar) - \tracer{\nabla \Loss(\Thetastar)}{\Delta} 
\le - \tracer{\nabla \Loss(\Thetastar)}{\Delta} 
+ \lambda \left ( \nucnorm{\Thetastar} - \nucnorm{\Thetahat}  \right )
\end{equation*}
By Taylor expansion, the left hand side is lower bounded by
\begin{equation*}
  \Loss(\Thetahat) - \Loss(\Thetastar) - \tracer{\nabla \Loss(\Thetastar)}{\Delta}  \ge \psi \left( 2\spiky \right ) \frac{1}{2\numobs} \sum_{i=1}^\numobs \left (\tracer{\Theta}{\design{i}} \right )^2
\end{equation*}
H\"older's inequality between the nuclear norm and operator norm yields
\begin{equation*}
- \tracer{\nabla \Loss(\Thetastar)}{\Delta} \le \matsnorm{\nabla \Loss(\Thetastar)}{2} \nucnorm{\Delta}
\end{equation*}
By the triangle inequality $\nucnorm{\Thetastar} - \nucnorm{\Thetahat}
\le \nucnorm{\Delta}$. If we choose $\lambda > 2\matsnorm{\nabla
  \Loss(\Thetastar)}{2}$, we have
\begin{equation*}
\Loss(\Thetahat) - \Loss(\Thetastar) - \tracer{\nabla \Loss(\Thetastar)}{\Delta} 
\le 2\lambda \nucnorm{\Delta} 
\end{equation*}
Now, the random matrix $\nabla \Loss(\Thetastar)=\frac{1}{\numobs}
\sum_{i=1}^{\numobs}\left (
  \frac{\exp(\tracer{\design{i}}{\Delta})}{1+\exp(\tracer{\design{i}}{\Delta})}
  - y_i \right ) \design{i}$ and satisfies the conditions of Lemma
\ref{Lem:crosstermbound} with $\gamma=2$, so we can take $\lambda=32
\sqrt{\frac{d \log d}{n}}$

From Lemma~1 of Negahban and Wainwright~\cite{NegWai11b}, $\Delta$ can be
decomposed into $\Delta'+\Delta''$, where $\Delta'$ has rank less than
$2r$ and $\Delta''$ satisfies
\begin{equation*}
\nucnorm{\Delta''}
\le 3\nucnorm{\Delta'} + 4 \sum_{j=r+1}^{\min\{\dima, \dimb\}} \sigma_j(\Thetastar)
\end{equation*}
Then by the triangle inequality and $\nucnorm{\Delta'} \le \sqrt{2r} \frobnorm{\Delta'}$
\begin{equation} 
\label{nucbound}
\nucnorm{\Delta}
\le 4\nucnorm{\Delta'} + 4 \sum_{j=r+1}^{\min\{\dima, \dimb\}} \sigma_j(\Thetastar)
\le 4\sqrt{2r} \frobnorm{\Delta} + 4 \sum_{j=r+1}^{\min\{\dima, \dimb\}} \sigma_j(\Thetastar)
\end{equation}

Now depending on whether $\Delta$ belongs to set $\setA$, we split into two cases. \\
\textbf{Case 1}: When $\Delta \notin \setA$, $\frobnorm{\Delta}^2 \le 128 \spiky \nucnorm{\Delta} \sqrt{\frac{\dimd \log \dimd}{n}}$. From Equation~\eqref{nucbound}, we get
\begin{equation*}
  \frobnorm{\Delta} \le \spiky\max \left \{1024 \sqrt{\frac{r\dimd \log \dimd}{n}}, \left( 512 \sqrt{\frac{r\dimd \log \dimd}{n}}\sum_{j=r+1}^{\min\{\dima, \dimb\}} \sigma_j(\Thetastar) \right )^{1/2} \right \}
\end{equation*}
\textbf{Case 2}: Otherwise, from Lemma \ref{RSC}, with probability greater than $1-2d^{-2^{18}}$,  $\Loss(\Thetahat) - \Loss(\Thetastar) -
\tracer{\nabla \Loss(\Thetastar)}{\Delta} \ge \frac{\psi \left( 2\spiky
\right )}{3} \frobnorm{\Delta}^2$. Therefore, the above equations yield
\begin{equation*}
  \frobnorm{\Delta}^2 \; \leq \frac{192}{\psi(2 \spiky)} \sqrt{\frac{2r \dimd \log \dimd}{\numobs}} \nucnorm{\Delta}.
\end{equation*}
Now, performing similar calculations as above we have
\begin{equation*}
\frobnorm{\Delta} \le \frac{1}{\psi(2 \spiky)} \max \left \{1024 \sqrt{\frac{r\dimd \log \dimd}{n}}, \left( 512 \sqrt{\frac{r\dimd \log \dimd}{n}}\sum_{j=r+1}^{\min\{\dima, \dimb\}} \sigma_j(\Thetastar) \right )^{1/2} \right \}.
\end{equation*}
Combining the two displays above yields the desired result.

\subsection{Proof of Lemma \ref{RSC}}
We use a peeling argument~\cite{geer2000} as in Lemma~3 of
\cite{NegWaiCompletion} to prove Lemma~\ref{RSC}. Before that, we
first present the following lemma.

\begin{lems}
  \label{deviations}
 	Define the set 
      \begin{equation*}
 	 \mathcal{B}(D) = \left \{ \Theta \in \reals^{\dima \times \dimb} \mid \infnorm{\Theta} \le r_3,  	
	 \frobnorm{\Theta} \le D,   \nucnorm{\Theta} \le \frac{D^2}{128\spiky} \sqrt{\frac{n}{d \log d}} \right \}
      \end{equation*}
	and 
      \begin{equation*}
         M(D) = \sup_{\Theta \in \mathcal{B}(D)} \left ( - \frac{1}{\numobs} \sum_{i=1}^\numobs \left (\tracer{\Theta}{\design{i}} \right )^2 + 2\frobnorm{\Theta}^2 \right )
      \end{equation*}
      Then 
      \begin{equation*}
      \prob \left \{  M(D) \ge \frac{3}{2}D^2 \right \} \le \exp \{ -\frac{nD^4}{128\spiky^4}\}
      \end{equation*}
\end{lems}

Since for any $\Theta \in \setA$, 
\begin{equation*}
\frobnorm{\Theta}^2 \ge  128 \spiky \sqrt{\frac{d \log d}{n}}\nucnorm{\Theta} \ge  128 \spiky \sqrt{\frac{d \log d}{n}}\frobnorm{\Theta} 
\end{equation*}
then we have $\frobnorm{\Theta} \ge 128 \spiky \sqrt{\frac{d \log d}{n}} := \mu$. Consider the sets
\begin{equation*}
\mathcal{S}_{\ell} = \left \{ \Theta \in \reals^{\dima \times \dimb} \mid \infnorm{\Theta} \le r_3,  	
	 \beta^{\ell-1} \mu \le \frobnorm{\Theta} \le \beta^\ell \mu,   \nucnorm{\Theta} \le \frac{D^2}{128\spiky} \sqrt{\frac{n}{d \log d}} \right \}
\end{equation*}
where $\beta=\sqrt{\frac{10}{9}}$ and $\ell=1,2,3\cdots$. \\
Suppose there exists $\Theta \in \setA$ such that $\frac{1}{\numobs} \sum_{i=1}^\numobs \left (\tracer{\Theta}{\design{i}} \right )^2 < \frac{1}{3}\frobnorm{\Theta}^2$. Since $\setA \subseteq\bigcup_{\ell=1}^{\infty} \mathcal{S}_{\ell}  \subseteq\bigcup_{\ell=1}^{\infty} \mathcal{B}(\beta^\ell \mu)$, there is some $\ell$ such that $\Theta \in \mathcal{B}(\beta^\ell \mu)$ and 
\begin{equation*}
- \frac{1}{\numobs} \sum_{i=1}^\numobs \left (\tracer{\Theta}{\design{i}} \right )^2 + 2\frobnorm{\Theta}^2 > \frac{5}{3} \frobnorm{\Theta}^2 \ge \frac{5}{3} \beta^{2\ell-2} \mu^2 =  \frac{3}{2} (\beta^{\ell} \mu)^2
\end{equation*}
Then by union bound, we have
{\setlength\arraycolsep{2pt}
\begin{eqnarray*}
&& \prob \left \{ \exists ~~\Theta \in \setA, ~\frac{1}{\numobs} \sum_{i=1}^\numobs \left (\tracer{\Theta}{\design{i}} \right )^2 < \frac{1}{3}\frobnorm{\Theta}^2 \right \} \\
&\le& \sum_{\ell=1}^{\infty} \prob \left \{ M(\beta^{\ell} \mu) > \frac{3}{2} (\beta^{\ell} \mu)^2 \right \}\\
&\le& \sum_{\ell=1}^{\infty} \exp \{ -\frac{n(\beta^{\ell} \mu)^4}{128\spiky^4}\} \\
&\le& \sum_{\ell=1}^{\infty} \exp \{ -\frac{ 4\ell (\beta-1) n\mu^4}{128\spiky^4}\} \\
&\le& 2 \exp \{ -\frac{ 4 (\beta-1) n\mu^4}{128\spiky^4}\} \\
&\le& 2 \exp \{ - 2^{18} \log d \}
\end{eqnarray*}}
where the second inequality is Lemma~\ref{deviations}, the third inequality is $\beta^\ell \ge \ell (\beta-1)$ and we use the fact that $n<d^2\log d$ for the last inequality.

\subsection{Proof of Lemma~\ref{deviations}}
Define
\begin{equation*}
  Z = : \frac{1}{\dima \dimb} M(D) = \sup_{\Theta \in \mathcal{B}(D)}\frac{1}{n} \sum_{i=1}^\numobs \left [ \E \Big(\Theta_{k(i)l(i)}-\Theta_{k(i)j(i)} \Big)^2 -\Big(\Theta_{k(i)l(i)}-\Theta_{k(i)j(i)} \Big)^2 \right ]
\end{equation*}

Our goal will be to first show that $Z$ concentrates around its mean
and then upper bound the expectation. We prove the concentration results via the
bounded differences inequality~\cite{ledoux2001}; since $Z$ is a symmetric function of its
arguments, it suffices to establish the bounded differences property
with respect to the first coordinate. Suppose we have two samples of
$(\useri{i}, \itema{i}, \itemb{i})_{i=1}^n$ that only differ at the
first coordinate.  
{\setlength\arraycolsep{2pt}
\begin{eqnarray*} 
Z- Z' 
% &=&\sup_{\Theta \in \setA} \frac{1}{n} \sum_{i=1}^{n} \Big(\Theta_{k'(i)l'(i)}-\Theta_{k'(i)j'(i)} \Big)^2 %-\sup_{\Theta \in \setA} \frac{1}{n} \sum_{i=1}^{n}  \Big(\Theta_{k(i)l(i)}-\Theta_{k(i)j(i)} \Big)^2 \\
&\le&  \sup_{\Theta \in \mathcal{B(D)}}  \Bigg[ \frac{1}{n}\sum_{i=1}^{n} (\Theta_{k'(i)l'(i)}-\Theta_{k'(i)j'(i)} \Big)^2 - \frac{1}{n}\sum_{i=1}^{n}  \Big(\Theta_{k(i)l(i)}-\Theta_{k(i)j(i)} \Big)^2 \Bigg ] \\
&=&  \sup_{\Theta \in \mathcal{B(D)}} \frac{1}{n} \Bigg( \Big(\Theta_{k'(1)l'(1)}-\Theta_{k'(1)j'(1)} \Big)^2 -   \Big(\Theta_{k(1)l(1)}-\Theta_{k(1)j(1)} \Big)^2 \Bigg) \\
&\le& \frac{4r_3^2}{n}
\end{eqnarray*}}
Then by the bounded differences inequality, we have 
\begin{equation} \label{bdiff}
\prob \{ Z - \E Z \ge t\} \le \exp\{ -\frac{nt^2}{32r_3^4}\}
\end{equation}

In order to upper bound $\E Z$, we use a standard symmetrization argument. 
{\setlength\arraycolsep{2pt}
\begin{eqnarray*}
\E Z &=& \E \sup_{\Theta \in \mathcal{B(D)}}\frac{1}{n} \sum_{i=1}^n \left [ \E \Big(\Theta_{a(i)l(i)}-\Theta_{a(i)j(i)} \Big)^2 -\Big(\Theta_{a(i)l(i)}-\Theta_{a(i)j(i)} \Big)^2 \right ] \\
&\le& \E \sup_{\Theta \in \mathcal{B(D)}} \frac{2}{n} \sum_{i=1}^{n} \varepsilon_i \Big(\Theta_{a(i)l(i)}-\Theta_{a(i)j(i)} \Big)^2  \\
&=& \E \sup_{\Theta \in \mathcal{B(D)}} \frac{2}{n} \sum_{i=1}^{n} \varepsilon_i \tracer{e_{k(i)} (e_{l(i)}-e_{j(i)})^T}{\Theta}^2
\end{eqnarray*}}
where $\varepsilon_i$ are i.i.d. Rademacher random variables.
Since $|\Theta_{a(i)l(i)}-\Theta_{a(i)j(i)}| \le 2r_3$, we have by the Ledoux-Talagrand contraction inequality that
\begin{equation*} 
\E \sup_{\Theta \in \mathcal{B(D)}} \frac{1}{n} \sum_{i=1}^{n} \varepsilon_i \tracer{e_{k(i)} (e_{l(i)}-e_{j(i)})^T}{\Theta}^2 \le 4 r_3 \E \sup_{\Theta \in \mathcal{B(D)}} \frac{1}{n} \sum_{i=1}^{n} \varepsilon_i\tracer{e_{k(i)} (e_{l(i)}-e_{j(i)})^T}{\Theta}
\end{equation*}
By an application of H\"older's inequality we have that
\begin{equation} |\sum_{i=1}^{n} \varepsilon_i \tracer{e_{k(i)} (e_{l(i)}-e_{j(i)})^T}{\Theta} |
\le \matsnorm{\sum_{i=1}^{n} \varepsilon_i  e_{k(i)} (e_{l(i)}-e_{j(i)})^T }2  \nucnorm{\Theta}
\end{equation}
Let $W_i := \varepsilon_i  e_{k(i)} (e_{l(i)}-e_{j(i)})^T$. $W_i$ is a zero-mean random matrix, and since 
\begin{equation*} 
\E [ W_i W_i^T] = \E [e_{k(i)} (e_{l(i)}-e_{j(i)})^T (e_{l(i)}-e_{j(i)}) e_{k(i)} ^T]= (2-\frac{2}{\dimb}) \frac{1}{\dima} \mathbf{I}_{\dima \times \dima} 
\end{equation*} 
and
\begin{equation*} 
\E [ W_i^T W_i] = \E [(e_{l(i)}-e_{j(i)}) e_{k(i)} ^T e_{k(i)} (e_{l(i)}-e_{j(i)})^T ]= \frac{2}{\dimb} \mathbf{I}_{\dimb \times \dimb} - \frac{2}{\dimb^2} \1\1^T 
\end{equation*}

we have
\[ \sigma_i^2 = \max \{\matsnorm{\E [ W_i^T W_i]}2,  \matsnorm{ \E [ W_i W_i^T] }2 \} \le\max \{ \frac{2}{\dimb},  (2-\frac{2}{\dimb}) \frac{1}{\dima} \} \le  \frac{2}{\min\{\dima,\dimb\}} \] 
Notice $\matsnorm{W_i}2\le 2$, thus, Lemma~\ref{Ahlswede-Winter} yields the tail bound 
\begin{equation}
\prob \Big[ \matsnorm{\frac{1}{n} \sum_{i=1}^{n} \varepsilon_i  e_{k(i)} (e_{l(i)}-e_{j(i)})^T}2 \ge t\Big] \le  \dima \dimb \max \{ \exp(-\frac{nt^2 \min\{\dima,\dimb\} }{8}), \exp(-\frac{nt}{4})\}
\end{equation} 
Set $t=\sqrt{\frac{16\log \dima\dimb}{n \min\{\dima,\dimb\}}}$, we obtain with probability greater that $1-\frac{1}{\dima \dimb}$, 
\begin{equation*}
\matsnorm{\frac{1}{n} \sum_{i=1}^{n} \varepsilon_i  e_{k(i)} (e_{l(i)}-e_{j(i)})^T}2 \le \sqrt{\frac{16\log \dima\dimb}{n \min\{\dima,\dimb\}}}
\end{equation*} 
By the triangle inequality, $\matsnorm{\frac{1}{n} \sum_{i=1}^{n} \varepsilon_i  e_{k(i)} (e_{l(i)}-e_{j(i)})^T}2 \le \matsnorm{\varepsilon_i  e_{k(i)} (e_{l(i)}-e_{j(i)})^T}2 \le 2$ and the fact $n \le d^2\log d$
\begin{equation}
\E \matsnorm{\frac{1}{n} \sum_{i=1}^{n} \varepsilon_i  e_{k(i)} (e_{l(i)}-e_{j(i)})^T}2 \le \sqrt{\frac{16\log \dima\dimb}{n \min\{\dima,\dimb\}}} + \frac{2}{\dima \dimb} \le 8 \sqrt{\frac{\log \dima\dimb}{n \min\{\dima,\dimb\}}} 
\end{equation} 
Putting those bounds together we have
\[ \E \sup_{\Theta \in \mathcal{B(D)}} \frac{1}{n} \sum_{i=1}^{n} \varepsilon_i \Big(\Theta_{a(i)l(i)}-\Theta_{a(i)j(i)} \Big)^2 \le \sup_{\Theta \in \mathcal{B(D)}} 32  r_3 \nucnorm{\Theta} \sqrt{\frac{\log \dima\dimb}{n \min\{\dima,\dimb\}}} \le \frac{D^2}{\dima \dimb} \] 
Plug it into \eqref{bdiff} and set $t = \frac{D^2}{2\dima \dimb}$, we get the result.

\subsection{Ahlswede-Winter Matrix Bound}
As in previous work~\cite{NegWaiCompletion} we also use a version of
the Ahlswede-Winter concentration bound. We use a version due to Tropp~\cite{randommatrix}.
\begin{lems}[Theorem~1.6~\cite{randommatrix}]
\label{Ahlswede-Winter}
  Let $W_i$ be independent $\dima \times \dimb$ zero-mean random matrices such that $\matsnorm{W_i}2 \le M$, and define
  \[ \sigma_i^2 := \max \{\matsnorm{\E [ W_i^T W_i]}2,  \matsnorm{\E [ W_i W_i^T] }2 \} \]
  as well as $\sigma^2 := \sum_{i=1}^{n} \sigma_i^2$. We have
  \begin{equation}
    \prob \Big[ \matsnorm{\sum_{i=1}^{n} W_i }2 \ge t\Big] \le  (\dima + \dimb) \max \{ \exp(-\frac{t^2}{4\sigma^2}), \exp(-\frac{t}{2M})\}
  \end{equation}
\end{lems}
\section{Discussion}
In this paper we presented a theoretical justification for a ranking
based collaborative filtering approach based on pairwise comparisons
in contrast to other results that rely on knowing the underlying
ratings. We provided the first convergence bounds for recovering the
underlying user preferences of items and showed that those bounds are
analogous to the ones originally developed for rating based matrix
completion. The analysis here can also be extended do other
observation models, for example to the ``one-bit'' matrix completion
setting as well. However, that extension does not provide any additional insights
beyond the analysis presented here. There remain a number of extensions for these methods
including adaptive and active recommendations, skewed sampling
distributions on the items, as well as different choice models. We
leave such extensions for future work.
\comment{\section{Acknowledgements}
The authors would like to thank Sewoong Oh and Devavrat Shah for many
helpful and inspiring conversations on rank aggregation.}

\bibliography{\bibdir/rankingbib}{}
\bibliographystyle{plain}

\end{document}